\documentclass[11pt,a4paper]{article}
\usepackage[hyperref]{emnlp-ijcnlp-2019}
\usepackage{times}
\usepackage{latexsym}

\usepackage{url}

\PassOptionsToPackage{usenames,dvipsnames,table}{xcolor}
\relax

\usepackage{helvet}  %Required
\usepackage{courier}  %Required
\usepackage{graphicx}  %Required
\usepackage{quoting}
\usepackage[utf8]{inputenc} % allow utf-8 input
\usepackage[T1]{fontenc}    % use 8-bit T1 fonts
\usepackage{booktabs}       % professional-quality tables
\usepackage{amsfonts}       % blackboard math symbols
\usepackage{nicefrac}       % compact symbols for 1/2, etc.
\usepackage{microtype}      % microtypography
\usepackage{amssymb}
\usepackage{amsmath}
\usepackage{multirow}
\usepackage{amsfonts}
\usepackage{booktabs}
\usepackage{rotating}
\usepackage{subfig}
\usepackage{array}
\usepackage{booktabs}
\usepackage{amsfonts}
\usepackage{enumitem}% http://ctan.org/pkg/enumitem
\usepackage{makecell}
\usepackage{xcolor}
\usepackage[toc,page]{appendix}
\usepackage{amsmath}
\usepackage{xspace}
\usepackage{mdwlist}
\usepackage{graphics}
\usepackage{color}
\usepackage{rotating}
\usepackage{booktabs}
\usepackage{epsfig}
\usepackage{alltt}
\usepackage{moreverb}
\usepackage{fancyvrb}
\usepackage{colortbl}      % for color table cells
\usepackage{xspace}
\usepackage{relsize}
\usepackage{array}
\usepackage{amsmath}
\usepackage{amsfonts}
\usepackage{txfonts}
\usepackage{caption}
\usepackage{rotating}
\usepackage[]{algorithm2e}
\usepackage{multirow}
\usepackage{booktabs}
\usepackage{float}
\usepackage{array,etoolbox}
\usepackage{enumitem} 
\usepackage{subfig}
\usepackage{arydshln}
\usepackage{setspace}
\usepackage{float}
\usepackage{mathtools}
\usepackage{bm}
\usepackage{xcolor}
\usepackage[toc,page]{appendix}
\usepackage{soul}
\usepackage[draft]{pgf}
\usepackage{makecell}
\usepackage[normalem]{ulem}
\usepackage{amssymb}
\usepackage{wasysym,textcomp}
\usepackage[toc,page]{appendix}
\usepackage{wrapfig}

\usepackage{tabularx}
\definecolor{Green}{rgb}{0,0.7,0}
\definecolor{Blue}{rgb}{0,0,1}
\definecolor{Orange}{rgb}{1,0.5,0}

\newcommand{\cameraready}[1]{{#1}}
\definecolor{cr}{rgb}{0.6,0,0}
\definecolor{Red}{rgb}{0,0,0}
\newcommand{\finaledit}[1]{\textcolor{Red}{#1}}
\usepackage{etoolbox}
\newtoggle{draft}
\toggletrue{draft}
\iftoggle{draft}{
\newcommand{\newresults}[1]{#1}
\newcommand{\todo}[1]{\textcolor{Red}{[\textsc{TODO: }#1 ]}}
}{
\newcommand{\todo}[1]{}
}

\newcommand{\statechange}[1]{\texttt{\textit{#1}}}
\newcommand{\entity}[1]{\texttt{#1}}

\newcommand{\com}[1]{}
\newcommand{\myparagraph}[1]{\vspace{1mm} \noindent {\bf #1: }}

\mathchardef\mhyphen="2D
\newenvironment{enu}{                   % list without par spacings
     \parskip 0cm \begin{list}{}{\parsep 0cm \itemsep 0cm \topsep 0cm}}{
       \end{list}} %  \parskip 0cm}

\newcommand{\prostruct}{\textsc{ProStruct}}
\newcommand{\ourmodel}{\textsc{XPAD}}
\newcommand{\ourmodelexpansion}{``eXPlaining Action Dependencies''}
\newenvironment{myquote}{                   % list without par spacings
  \parskip 0mm \begin{quoting}[vskip=0mm,leftmargin=2mm]}{
\end{quoting}}

\newcommand{\dataset}{ProPara}

\aclfinalcopy % Uncomment this line for the final submission
\title{Everything Happens for a Reason: \\ 
Discovering the Purpose of Actions in Procedural Text}
\author{ 
Bhavana Dalvi Mishra\textsuperscript{*},\quad  Niket Tandon\thanks{\textsuperscript{*}Bhavana Dalvi Mishra and Niket Tandon contributed equally to this work.},
\quad Antoine Bosselut, \\ 
        {\bf Wen-tau Yih, \footnotemark}
        \quad {\bf Peter Clark}  \\
        Allen Institute for Artificial Intelligence, Seattle, WA \\
        {\tt \{bhavanad, nikett, antoineb, peterc\}@allenai.org}
        }
        
\date{}

\hypersetup{final}
\begin{document}
\maketitle

\begin{abstract}
Our goal is to better comprehend procedural text, e.g., a paragraph about photosynthesis, by not only predicting what happens, but {\it why}  some actions need to happen before others.  Our approach builds on a prior process comprehension framework for predicting actions' effects, to also identify subsequent steps that those effects enable. 
We present our new model (\ourmodel) that biases effect predictions towards those that (1) explain more of the actions in the paragraph and (2) are more plausible with respect to background knowledge. 
\finaledit{We also extend an existing benchmark dataset for procedural text comprehension, ProPara, by
  adding the new task of explaining actions by predicting their dependencies. We find that \ourmodel~significantly
  outperforms prior systems on this task, while maintaining the performance on the original task in ProPara.}
\cameraready{The dataset is available at \small \url{http://data.allenai.org/propara}}

\end{abstract}

\footnotetext[1]{Scott is currently at Facebook AI Research (scottyih@fb.com)}

\section{Introduction}

\begin{figure}[t]
% \centerline{
\begin{center}
{\includegraphics[width=\columnwidth]{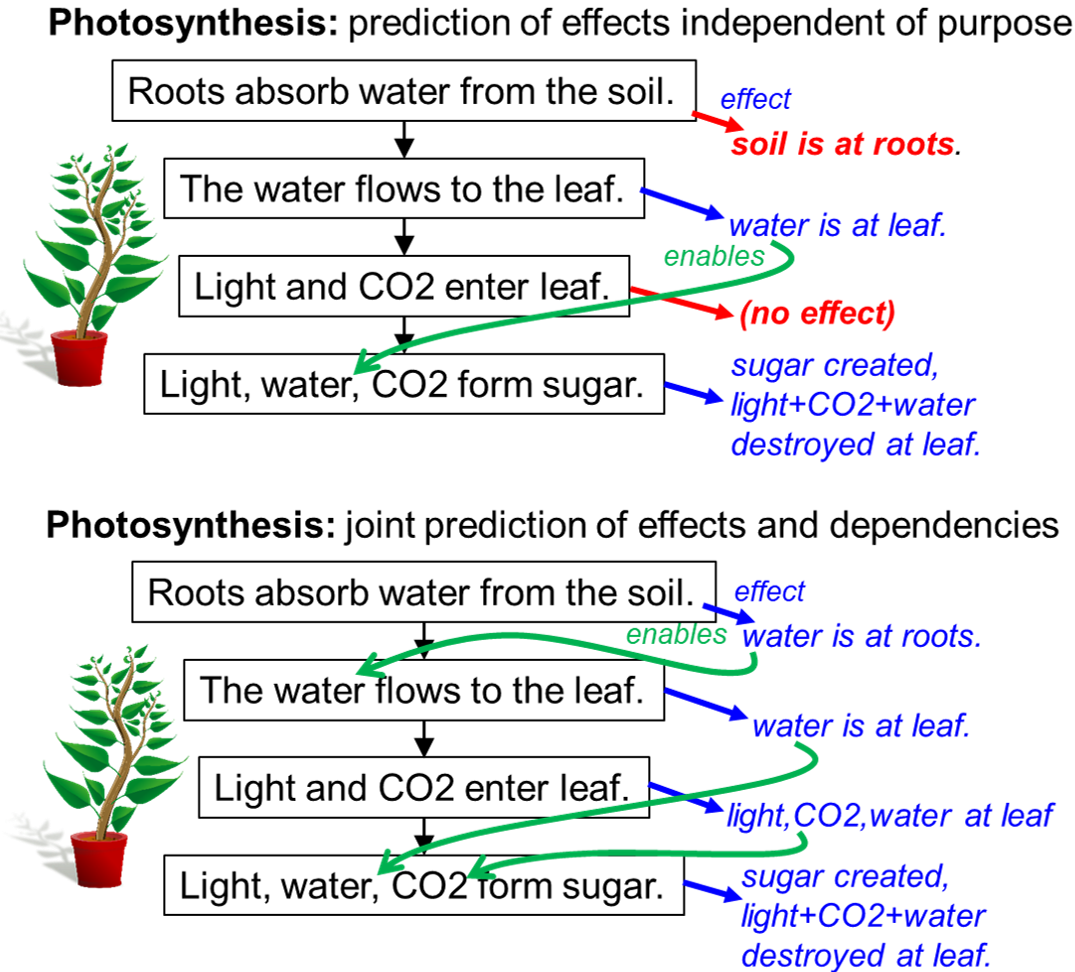}}
\end{center}
% }
\caption{
An action sequence (black) describes the order that actions happen, but not why.
We add the new task of also predicting {\it why} the actions are needed, in the form of the actions' effects (blue) and subsequent actions that depend on those effects (green).
While a system that predicts effects without considering dependencies can make errors (red, top part), we obtain 
% significantly 
better predictions by biasing the system to predict effects that \finaledit{{\it also}} result in more dependencies (lower part).
}
\label{fig:example}
    \vspace{-2mm}
\end{figure}

Procedural text is common in natural language, for example in recipes, how-to guides,
and science processes, but understanding it remains a major challenge.
While there has been substantial prior work on extracting
the sequence of actions (``scripts'') from such texts,
the task of identifying {\it why} the sequence is the way it 
is has received less attention, and is the goal of this work.
While ``why'' can mean many things, we treat it here as
describing how one action produces effects that are required by another.
For example in Figure~\ref{fig:example}, ``CO2 enters the leaf'' is necessary
because it results in ``CO2 is at the leaf'', a precondition for ``CO2 forms sugar''.
% (An important insight is that 
(Note that 
an action does not directly depend on previous actions, but
rather on the {\it state of the world} resulting from previous actions).
If one could determine such rationales, new capabilities would become possible, including explanation,
identifying alternative event orderings, and answering "\emph{what if...}" questions.
However, this task is challenging as it requires knowledge of
the preconditions and effects of actions, typically unstated in the text itself.

Recent work in neural process modeling goes partway towards our goal
by modeling the effects of actions, using annotated training data, allowing the
states of entities to be tracked throughout a paragraph,
e.g., EntNet~\cite{Henaff2016TrackingTW}, NPN~\cite{npn}, and ProStruct~\cite{propara-emnlp18}.
However, these systems do not consider the purpose of those effects in
the overall action sequence. As a result, they are unaware if they predict
effects that have no apparent purpose in the process, possibly
indicating a prediction error (e.g., the erroneous predictions
in red in Figure~\ref{fig:example}).

To address these limitations, we extend procedural text comprehension with an
additional task of predicting the dependencies between steps, in the form of
their effects and which subsequent action(s) become possible.
Building upon the state-of-the-art framework for predicting effects of actions, we present a new model, called \ourmodel~(\ourmodelexpansion) that also considers the purpose of those effects.
Specifically, \ourmodel~biases those predictions towards
those that (1) explain more of the actions in the paragraph and (2) are more
plausible with respect to background knowledge. On a benchmark dataset for
procedural text comprehension, ProPara \cite{propara-naacl18},  \ourmodel~significantly improves on
the prediction and explanation of action dependencies compared to prior systems, while
also matching state-of-the-art results on the original tasks.
We thus contribute:
\begin{enu}
\item[1.] A new task for procedural text comprehension, namely predicting and explaining the dependencies between actions (``what depends on what, and why''), \finaledit{including an additional dependency graph dataset for ProPara.}
\item[2.] A model, \ourmodel, that significantly outperforms prior systems at predicting and explaining action dependencies,  while maintaining its performance on the original tasks in ProPara.
\end{enu}

\begin{figure*}[t]
\begin{center}
  {\includegraphics[width=2\columnwidth]{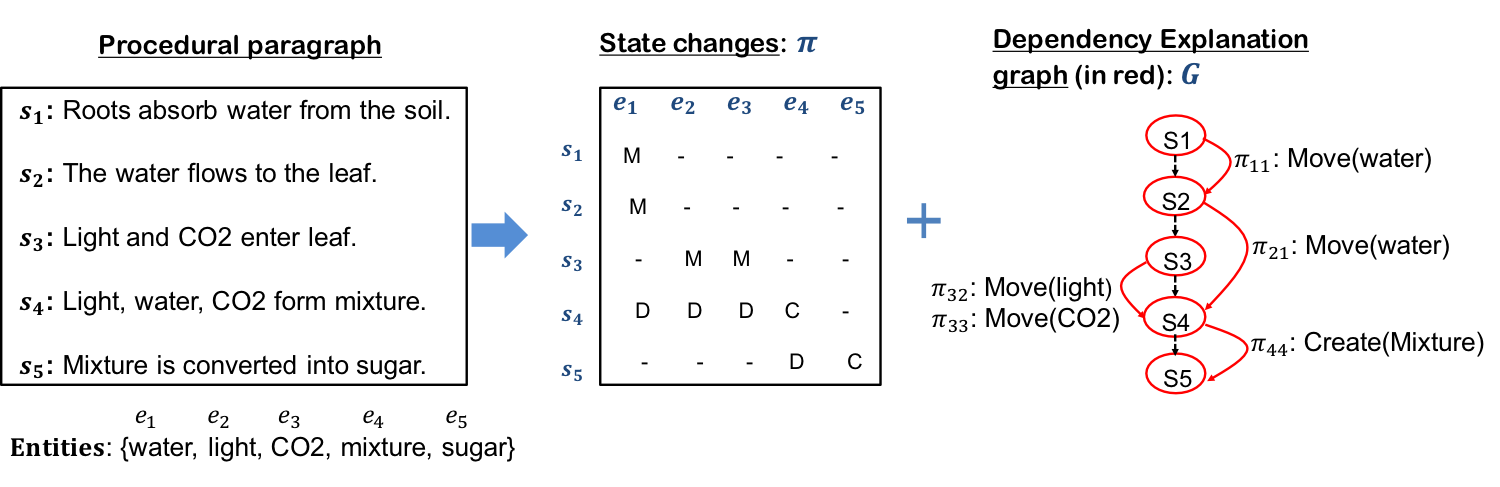}}
\vspace{-3mm}
\end{center}
%}
\caption{A procedural paragraph and target entities are provided as input. The task is to predict the state change matrix $\pi$, along with the dependency explanation graph $\mathcal{G}$ between steps (red, overlayed on the step sequence). $\pi_{tj}$ refers to the state change happening to entity $e_j$ in step $s_t$. In the ProPara task, there are 4 possible state changes denoted by 'M' (Move), 'C' (Create), 'D' (Destroy), and '-' (no change). 
Further, $\mathcal{G}$ describes which steps enable which other steps and how. For example, 
$s_4$ enables $s_5$ by creating \entity{mixture}, required by $s_5$.
  }
  \vspace{-3mm}
\label{fig:io_example}
\end{figure*}

\section{Related Work}

Understanding procedural text has a long history in AI. Early work
attempted to construct semantic representations (``scripts'') of
event sequences (or partial orders) from text, including representations
of the goals, effects, and purpose of actions,
e.g., \cite{schank1977scripts,dejong1979prediction,cullingford1986sam,mooney1986generalizing}.
Some of these systems could explain why actions occurred in text,
similar to our goals here, but only on a handful of toy examples
using hand-coded background knowledge, and proved difficult to scale.
More recent work on event extraction and script learning
has proved more effective at extracting event/action sequences from text
using statistical methods, e.g., \cite{Chambers2008UnsupervisedLO},
and neural techniques, e.g., \cite{Modi2014InducingNM,Modi2016EventEF,Pichotta2016UsingSL},
but have largely focused on {\it what} happens and in what order, rather than {\it why}.
For example, such representations cannot answer questions about which events 
would fail if an earlier action in a sequence did not occur.
The 2014 ProRead system \cite{Scaria2013LearningBP,berant2014modeling}
included dependency relationships between events that it extracted,
but assessed dependencies based on surface language cues,
hence could not explain why those dependencies held.

There has also been a line of research in reading procedural text
in which the goal has been to track how entities' states change
with time, e.g., EntNet \cite{Henaff2016TrackingTW},
ProStruct \cite{propara-emnlp18},
and Neural Process Networks \cite{npn},
applied to procedural text such as cooking recipes \cite{Kiddon2016GloballyCT},
science processes \cite{propara-naacl18}, or toy stories \cite{weston2015towards}.
Using annotated data, these systems learn to predict the effects of
actions from text, allowing simple simulations of how the world changes
throughout the process. We build on this line of work to identify
the purpose of actions, by connecting their effects to subsequent
actions.

In addition to signal from training data, many systems use
background knowledge to help bias predictions towards those
consistent with that knowledge.
For example, \cite{McLauchlan2004ThesaurusesFP} predicts
prepositional phrase attachments that are more consistent
with unambiguous attachments derived from thesaurii; 
\cite{Clark2009LargescaleEA} tune a parser to prefer bracketings
consistent with frequent bracketings stored a text-derived
corpus of bracketings; and \cite{propara-emnlp18} predicts state changes
consistent with those seen in a large text corpus. 
For our purposes, while KBs such as \cite{Speer2013ConceptNet5A, HowToKB, WikiHowNeural2018} contain useful information about action dependencies (e.g., ``smoking can cause cancer''), they lack explanations for those links. Instead, we create a KB of dependency links with explanations (Section~\ref{background-knowledge}), allowing us to similarly bias predictions with background knowledge.

\section{Problem Definition}
\label{sec:problem-definition}
%\todo{SY: We need to properly mention that this problem is derived from the ProPara task...}
The input to our system is a paragraph of procedural text, along with a list of the \emph{participant entities} in
the procedure. The output is a \emph{state change matrix} highlighting the changes undergone by the entities, as well as a 
\emph{dependency explanation graph} that encodes ``what action\footnote{
We use a broad definition of action to mean any event that
changes the state of the world (including non-volitional events
such as roots absorbing water).
} depends on what, and why'', as illustrated in
Figure~\ref{fig:io_example}. This problem definition is derived from the one used for the
ProPara dataset \cite{propara-naacl18}, with the key addition of the dependency explanation graph.
Although we use ProPara to illustrate and evaluate our work here, our approach is not specific to the
details of that particular dataset.
%(discussed in Section~\ref{discussion}).

% We give the formal definition of these components below.
% We formally define these now.

\myparagraph{Paragraph}
We define a paragraph of procedural text $S = \langle s_1, \cdots, s_T \rangle$ as a sequence of sentences that describe a procedure (e.g., photosynthesis).
We assume that sentences are chronologically ordered and treat a sentence $s_t$ as a step executed at time $t$.\footnote{This assumption holds in ProPara, the dataset used in this work. However, our techniques can generally be applied as long as a partial order of events is given. 
% (Section~\ref{discussion}).
}

\myparagraph{Participant Entities}
We  assume that the set of participant entities $E = \{e_1, \cdots, e_n\}$ in the paragraph $S$ 
is given. 
%Each participant entity is denoted by the set of its mentions.
Notice that $E$ 
includes only entities that participate in the process where their state changes at some steps.

\myparagraph{State Change Matrix}
For each entity $e_j \in E$, the system tracks how its properties change (e.g., location, existence) after each step.
We enumerate all the state changes in a $T \times n$ matrix $\pi$, where $\pi_{tj}$ denote how $e_j$ has changed
after step $s_t$. For ProPara, we use just four possible state changes
for $\pi_{tj}$, namely \{\statechange{Move, Create, Destroy, None}\}, sufficient
to model the properties tracked in the dataset. (More generally, additional state changes could be used provided their
effects on entity properties is clearly defined.)
\statechange{Move} means that the location of $e_j$ changes after $s_t$.
\statechange{Create} and \statechange{Destroy} indicate $e_j$ starts or ceases to exist.
\statechange{None} means that the state of $e_j$ remains unchanged. % after $s_t$.
A state change can optionally take arguments, e.g., \statechange{Move} is associated with before and after locations.

\myparagraph{Dependency Explanation Graph}
In addition to being able to predict when entities undergo state changes, the system also needs to understand the dependency relationships 
between steps in a procedure. That is, for a step $s_i$ to be executable, the system must be able to identify which previous steps are required to have been completed first.
We represent these dependency relationships in procedural text using a \emph{dependency explanation graph} $\mathcal{G} = \langle S, \mathcal{E} \rangle$, where each \textbf{node} is a step $s_i \in S$
and a \textbf{directed edge} $(s_i,s_j) \in \mathcal{E}$ indicates that $s_i$ is a precondition of $s_j$ ($i < j$).
% , as we assume the steps in $S$ are chronologically ordered). 
Moreover, each edge is associated with the \textbf{explanation} in the form of the entity state change that begets this dependency.
For example, in the example shown in Fig.~\ref{fig:io_example}, both $s_2$ and $s_3$ are the parent nodes of $s_4$.
For \emph{light, water, CO$_2$} to form the \emph{mixture} in $s_4$, the \entity{water} has to move to the leaf in $s_2$ (i.e., $\pi_{21} =~$\statechange{Move}), 
and \entity{light} and \entity{CO$_2$} also need to come to the leaf in $s_3$ (i.e., $\pi_{32} = \pi_{33} =~$\statechange{Move}). 
Note that the dependency graph is not the ``script'' (event sequence) of the process, but an overlay on the
script that explains {\it why} some actions need to happen before others. 
\cameraready{Note also that the state changes are fully specified (including the from and to locations) - the location information is dropped in Figure~\ref{fig:io_example} for simplicity.}

% \paragraph{Computing Dependency Explanation Graph $\mathcal{G}$ }

\section{The Dependency Graph Dataset}

% \paragraph{Creating Dependency Explanation Graphs $\mathcal{G}$:}
\label{dependency-graph-computation}

\finaledit{
Dependency graphs were added to the ProPara dataset using a mixture of manual and automated methods.
First, an algorithm was used to estimate dependencies using a heuristic method (below).
Then, for the test set, these dependencies were manually reviewed and corrected.
Approximately 15\% of these dependencies needed to be changed,
suggesting the heuristic algorithm is a reasonable approximation of the ground truth.
The train and dev sets remain with the automatically generated (noisy but more numerous) dependency graphs.}

\finaledit{As well as helping to augment the ProPara dataset, the algorithm can be used to automatically
add approximate explanation graphs to existing models' state change predictions. On first sight,
this might seem like a simple solution to the dependency prediction task. However, as we show
later, this approach does not produce good results. This is because models' state
change predictions were generated without regard for their explainability, and
hence prediction errors can cascade into explainability errors. A better solution,
which is the basis of XPAD, is to minimize a {\it joint} loss for both state change
and explainability. In this way, an algorithm will be steered towards state
changes that are also explainable (Figure~\ref{fig:example}).}

\finaledit{The dependency graph algorithm is as follows, and is based on a {\it coherence assumption:}}
If step $s_j$ changes the state of entity $e_k$,
we assume that the reason $s_j$ was included in the paragraph is because the next step mentioning $e_k$ requires that
change as a precondition. 
By searching forward for the
first subsequent step $s_j$ in which $e_k$ is again mentioned, \finaledit{or changes state again,} we add  
an enable edge in the dependency explanation graph that points from $s_i$ to $s_j$, with the explanation label $\pi_{ik}$.
As shown in Sec.~\ref{dependency-graph-errors}, these dependency graphs that we added  are mostly accurate (F1 $\approx$ 0.85).

\section{\ourmodel{} Model} 
\label{sec:procausal}

Our model builds on the approach used in the ProStruct system \cite{propara-emnlp18},
namely an encoder-decoder approach with beam search decoding.
We follow its design for the encoder, but use a modified
decoder that also generates dependency graphs, and biases search towards graphs that
are both more connected and more a priori likely.

\begin{figure}[t]
\begin{center}
{\includegraphics[width=\columnwidth]{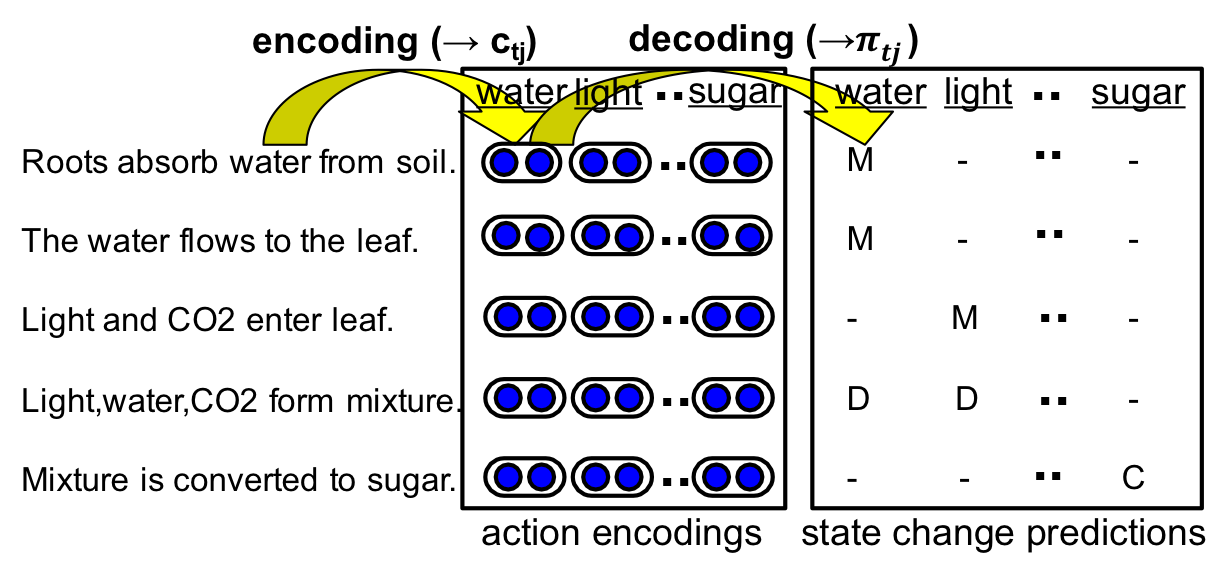}}
\end{center}
\vspace{-1em}
\caption{Encoder-decoder architecture used in \ourmodel: First encode the action of each sentence $s_t$ on entity $e_j$, then decode to a predicted state change (state change arguments not shown). 
For example, the action \emph{Roots absorb water from soil} is predicted to \statechange{Move} (M) the \entity{water} (from soil to roots, not shown). Global predictions are generated through beam search.}
\label{encoder-decoder}
\vspace{-3mm}
\end{figure}

\begin{figure}[t]
\begin{center}
{\includegraphics[width=0.70\columnwidth]{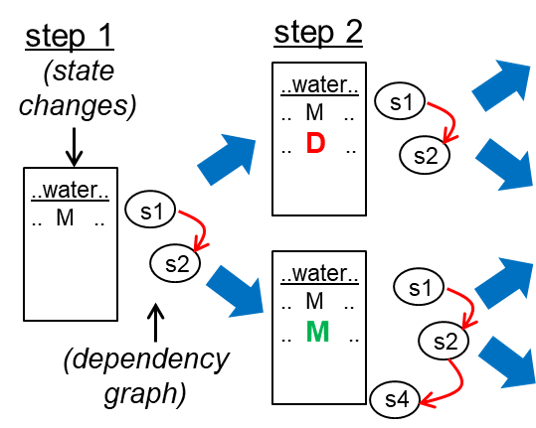}}
\end{center}
\caption{\ourmodel's beam search for the best decoding, one sentence (step) at a time, showing the predicted state changes and the resulting dependency graph after each step. In step 2, \ourmodel~chooses between predicting whether \entity{water} is \statechange{destroyed} (\textcolor{red}{\bf D}, upper figure) or \statechange{moved} (\textcolor{green}{\bf M}, lower). Predicting \statechange{M} results in a more connected dependency graph (as water is mentioned again in step 4), hence the lower choice is likely preferred (Eq.~\eqref{eq:h-graph}) by our \emph{\textbf{dependency graph score}}. This score also 
assesses how a priori likely it is that $s_4$'s movement of water enables $s_5$, using a background KB (Eq.~\eqref{eq:h_kb}).}
\label{beam-search}
\vspace{-3mm}
\end{figure}

% \vspace{-1em}
\subsection{Encoder}
For each sentence $s_t$ and each entity $e_j$, the encoder creates a vector $c_{tj}$, capturing how the actions in $s_t$ affect $e_j$ (Figure~\ref{encoder-decoder}).
During encoding, each word $w_i$ in $s_t$ is first
represented by its GloVe vector, concatenated with two indicator variables:
whether the word refers to $e_j$ and whether the word is a verb.
The 
%whole 
sentence is then fed into a BiLSTM to generate a contextualized vector $h_i$,
which is then passed to a bilinear attention layer:
%\begin{align}
%    a_i &= h_i \mathrm{B} h_{ev} + b,
%\end{align}
$ a_i = h_i \mathrm{B} h_{ev} + b$,
where $B$ and $b$ are learned parameters, and $h_{ev}$ is the concatenation
of $h_e$ and $h_v$ (the averaged contextualized embedding of the entity and the verb words, respectively).
The output vector $c_{tj}$ is the attention-weighted sum of the $h_i$:
$ c_{tj} = \sum_{i=1}^I a_{i} \cdot h_i$.

Each $c_{tj}$ will be decoded (Sec.~\ref{sec:decoder}) into one of $K$ possible state {\it changes}. % In this work, 
(For the ProPara application, $K=4$ for \{\statechange{create}, \statechange{move}, \statechange{destroy},  \statechange{none}\}).
As preparation for this, we pass each $c_{tj}$ through a feedforward layer to generate $K$ logits that denote how likely each entity $e_j$ is to undergo each state change during step $s_t$.
For changes that take additional arguments (e.g., \statechange{move}), the arguments are
chosen using standard span prediction techniques
over the encoded input $h_i$~\cite{Seo2016BidirectionalAF}.

\subsection{Decoder \label{sec:decoder}}

The goal of decoding is to determine the state change matrix $\pi$ and the dependency graph $\mathcal{G}$.
While each element $\pi_{tj}$ of the state change matrix $\pi$ could be determined solely by choosing
the state change with the highest logit value, this greedy, \emph{local} approach
often results in nonsensical predictions (e.g., an entity is moved before being created).
To avoid this, \ourmodel~instead performs a beam search of possible decodings, 
one sentence $s_t$ at a time, using a scoring function that includes terms that
ensure {\it global} consistency, as shown in Fig.~\ref{beam-search}.
Let $\pi_{t}$ be the state change matrix up to step $t$, and $\mathcal{G}_t$ be the
dependency graph up to step $t$. As $\pi_{t}$ is constructed,
$\mathcal{G}_t$ is derived from it deterministically using the same
heuristic procedure described in Section~\ref{dependency-graph-computation}.
Most importantly, the scoring function used in the beam search is a
function of {\it both} $\pi_t$ and $\mathcal{G}_t$.  This directs search
towards state change predictions that are both globally consistent and
produce a dependency graph that is well connected and a priori likely.
Finally, the decoder outputs the complete state change grid $\pi~(= \pi_T)$ and 
dependency graph $\mathcal{G}~(= \mathcal{G}_T)$ from step $1$ to $T$.

\subsection{Dependency Aware Scoring Function}
% Define scoring function: likelihood of action sequence, coherence of dependency graph
At each decoding step $t$, \ourmodel{} explores different options of $\pi_t$ (state changes for all entities) and $\mathcal{G}_t$ (corresponding dependency explanation graphs) as illustrated in Fig.~\ref{beam-search}.
In particular, it scores the candidates at step $t$ based on two components:
\begin{align}
\label{eq:procausal_score}
% score = state-change-score + dependency-graph-score
\phi(\pi_{t}, \mathcal{G}_t) &= \lambda \cdot f(\pi_{t}) + (1 - \lambda) \cdot g(\mathcal{G}_t | \pi_{t}),
\end{align}
where (1) $f(\pi_{t})$ is the \emph{\textbf{state change score}} based on the likelihood of selected state changes at step $t$ given the text and the state change history from steps $s_1$ to $s_{t-1}$, (2) $g(\mathcal{G}_t | \pi_{t})$ is the \emph{\textbf{dependency graph score}} based on the connectivity and likelihood of the resulting dependency explanation graph $\mathcal{G}_t$, and (3) $\lambda$ is a hyper-parameter that determines the importance of the accuracy of the state changes vs. the coherence of the action dependency graph.
%

% Define f(pi_t)
\paragraph{State Change Score:}
To compute $f(\pi_{t})$, we reuse the scoring function proposed by ProStruct:
% The scoring function for evaluating which $\pi_t$ to add to the current sequence $\pi$
% %consists of 
% has
% two terms: the %original 
% logit value for the local prediction and a bias term
% reflecting \emph{a priori} likelihood of the state changes:
\begin{equation}
\small{
  f(\pi_{t}) = \sum_{j=1}^{|E|} \Big( \alpha \cdot \mathrm{logit}(\pi_{tj}) + (1 - \alpha) \cdot \text{ }  \log P_{\mathrm{ext}}(\pi_{tj} | e_j) \Big),
\label{eq:kb}
}
\end{equation}
where
$\alpha$ is a hyper-parameter controlling the degree of bias,
$\mathrm{logit}(\pi_{tj})$ is the logit value supplied by the encoder for the local prediction, and
 $P_{\mathrm{ext}}(\pi_{tj} | e_j)$ indicates how likely entity $e_j$ will go through
some change defined in $\pi_{tj}$, based on the topic of the procedural text.
% , as observed in an external corpus. 
We use the knowledge base published by \citet{propara-emnlp18} to compute  $P_{\mathrm{ext}}(\pi_{tj} | e_j)$.

\paragraph{Dependency Graph Score:}
$g(\mathcal{G}_t | \pi_{t})$ is the score of the resulting dependency graph given the selected state changes $\pi_{t}$, defined as:
\begin{align}
\small{
\label{eq:procausal_components}
g(\mathcal{G}_t | \pi_{t}) = \beta \cdot g_{edge}(\mathcal{G}_t | \pi_{t}) + (1 - \beta) \cdot g_{kb}(\mathcal{G}_t)
}
\end{align}
where (1) $g_{edge}(\mathcal{G}_t | \pi_{t})$ scores $\pi_{t}$'s contributions to the dependency graph,
(2) $g_{kb}(\mathcal{G}_t)$ encodes how \emph{a priori} likely the added dependency edges, computed using background knowledge, and (3) $\beta$ is a hyperparameter tuned on the dev set.  

With the help of the \emph{\textbf{dependency graph score}},  \ourmodel{} is able to bias the search toward predictions that 
have an identifiable purpose (i.e., dependency edges), and toward dependency graphs that are more likely according to the background knowledge. 
Next, we describe scores $g_{edge}$ and $g_{kb}$ in more details.
%%---------------------------------------------------------------------------------------------------------

\paragraph{(a) Score $g_{edge}$: Actions with Purpose \\ }
\label{bias1}

% We use the dependency graph $DG$ to 
We compute $g_{edge}$ in Eq.~\eqref{eq:procausal_components} to bias the search toward dependency graphs %($\mathcal{G}$'s) 
that are more connected:
\begin{equation}
\label{eq:h-graph}
g_{edge}(\mathcal{G}_t | \pi_{t}) = \sum_{j=1}^{\vert E \vert} \log\Big( 1 + \mathrm{score}_{p} (\mathcal{G}_t |\pi_{tj}) \Big), 
\end{equation}
where $score_{p}$ is assigned based on whether $\pi_{tj}$ results in a purpose for step $s_t$ or not, as follows:
\begin{align*}
\label{eq:purpose_score}
\mathrm{score}_p (\mathcal{G}_t | \pi_{tj})  = 
\begin{cases}
       +c, \quad\text{if $\pi_{tj}$ adds an edge to $\mathcal{G}_t$;}\\
       0, \quad\text{if no edge can be added  }\\
       \quad\text{\ \ to $\mathcal{G}_t$ as $e_j$ isn't mentioned later;}\\
        -c, \quad\text{if $\pi_{tj}$ doesn't add an edge }\\
            \quad\text{\ \  to $\mathcal{G}_t$ and $e_j$ is mentioned later.}
     \end{cases}
\end{align*}
where $0 < c < 1$ is a hyper-parameter tuned on the dev set.
Conceptually, this scoring function
% The $\mathrm{score}$ function essentially 
checks whether a particular state change prediction results in a connection in the dependency explanation graph $\mathcal{G}_t$.
For example, if $\pi_{tj}$ = \statechange{Create entity $e$} and $e$ is mentioned in a later step $s_k$, then an  edge will be added between $s_t$ and $s_k$ to $\mathcal{G}_t$. Consequently,  $\mathrm{score}_p = c > 0$, resulting in a positive boost in the score for $\pi_{tj}$.

%%------------------------------------------

%%-------------------------------------------

\paragraph{(b) Score $g_{kb}$: Action Dependency Priors\\}   
\label{background-knowledge}

To distinguish likely dependency links (e.g., $\mathcal{E}_{ij}$ in $\mathcal{G}_t$) from unlikely ones, we also bias the predictions in Equation~\ref{eq:procausal_components} with $g_{kb}$. 
The scoring function $g_{kb}$ computes the prior likelihood of each dependency link $\mathcal{E}_{ij}$ that would be added to $\mathcal{G}_t$, as a result of the predicted actions $\pi_{ik} \in \pi_{i}$. This bias is particularly helpful when there is a limited amount of training data.
If $\mathcal{E}_t$ is the set of edges added to $\mathcal{G}_t$ because of selected state changes $\pi_{t}$, then
\begin{equation}
\label{eq:h_kb}
    g_{KB}(\mathcal{G}_t) = \sum_{\mathcal{E}_{ij} \in \mathcal{E}_t} log\Big( 1 + score_{e}(\mathcal{E}_{ij}) \Big)
\end{equation}
where $score_{e}(\mathcal{E}_{ij})$ scores the action dependency link (if any, 0 otherwise) added by action $\pi_{ik}$ on an entity $e_k$ that is common to steps $s_i$ and $s_j$.
For example, in Figure~\ref{fig:io_example}, action $\pi_{11}$ adds an edge \statechange{MOVE(water)}, between sentences ``\emph{Roots absorb water.}" and ``\emph{The water flows to the leaf.}" 

We need to estimate the likelihood of $\mathcal{E}_{ij}$ (i.e., the likelihood that $s_i$ can enable $s_j$ via $\pi_{ik}$). This information (especially $\pi_{ik}$) is not present in any existing KB. Therefore, we train a model using a large collection of positive and negative examples of valid/invalid edges. 

To generate these training examples, we first extract a large collection of procedural texts from WikiHow.com, which contains 1.75M steps from 227K processes across real-world domains, including health, finance, education, home, food, hobbies, etc.
On each text, we apply a rule-based system %called ProComp 
\cite{propara-arxiv} that noisily generates $\pi_{train}$, and then apply the heuristics from Section~\ref{dependency-graph-computation} to obtain $\mathcal{G}_{train}$. The distribution in $\mathcal{G}_{train}$ can be quite different from that in ProPara (the current task), so we append $\mathcal{G}_{train}$ with dependency graphs obtained from the training set in ProPara. We then decompose $\mathcal{G}_{train}$ into its $\mathcal{E}_{ij}$ edges (negative examples are created by reversing these edges). This leads to 324,462 training examples. 
We then add 2,201 examples derived from the ProPara train set.
This use of hand-written rules to generate a large number of potentially noisy examples follows others in the literature,
e.g. \cite{Sharp2016CreatingCE,ahn2016neural,npn}.

To accommodate for lexical variations, we embed the database of training data in a neural model. 
Our model itself takes as input $\mathcal{E}_{ij}$ and outputs the likelihood of $\mathcal{E}_{ij}$. To do this, an embedding for $\mathcal{E}_{ij}$ is created using a deep network of biLSTMs, producing a contextual embedding based on the token level embeddings of $s_i, s_j$ and the state change vector $\pi_{ik}$. This contextual embedding is then decoded using a feedforward network to predict a score for whether $s_i$ enabled $s_j$ through $\pi_{ik}$. The loss function is designed such that errors on the training examples coming from ProPara are penalized $\theta$ times more than those from WikiHow, where $\theta$ is a hyperparameter tuned on the dev set. 

\cameraready{While $g_{edge}$ only scores whether the same entity is used in the two connected events or not,
ignoring location information, $g_{kb}$ models how likely there is a dependency between
the two events, including using the from/to location of an entity. For example,
$g_{kb}$ can model that moving `light' and `CO2' to `location=leaf' is important before 
they can be turned into a `mixture' (Figure~\ref{fig:io_example}).}

\subsection{Training and Testing \ourmodel}
% \paragraph{Training and Testing \ourmodel:}
At training time, \ourmodel{} follows only the correct (gold) path through the search space, and learns to minimize
the joint loss of predicting the correct state changes and dependency explanation graph for the paragraph.
At test time, \ourmodel~performs a beam search to predict the most likely state changes and dependency explanation graph. %\footnote{The data and model code is available at \small \url{http://data.allenai.org/propara}.}.

\subsection{Implementation Details for \ourmodel}
We implement \ourmodel{} in PyTorch using AllenNLP \cite{allennlp}. We use the dataset reader published in ProStruct's publicly available code. We use 100D Glove embeddings \cite{pennington2014glove}, trained on Wikipedia 2014 and Gigaword 5 corpora (6B tokens, 400K vocab, uncased). Starting from glove embeddings appended by entity and verb indicators, we use bidirectional LSTM layer to create contextual representation for every word. We use 100D hidden representations for the bidirectional LSTM~\cite{hochreiter1997long} shared between all inputs (each direction uses 50D hidden vectors). The attention layer on top of BiLSTM, uses a bilinear similarity function similar to~\cite{Chen2016ATE} to compute attention weights over the contextual embedding for every word. 

To compute the likelihood of all state changes individually, we use a single layer feedforward network with input dimension of 100 and output 4. We then use constrained decoder during training as explained in Section \ref{sec:decoder}. We tune the hyperparameters in Equations 1, 2, and 3 as $\lambda=0.5$, $\alpha=0.8$,  $\beta=0.8$. We use Adadelta optimizer \cite{zeiler12} with learning rate 0.2 to minimize total loss.

To make predictions, we use a constrained decoder with beam size of 20, i.e. top 20 choices are explored at each step.% in the paragraph.

%%%%%%%%%%%%%%%%%%%%%%%%%%%%%%%%%%%%%%%%%%%%%%%%%%%%%%%%%%%%%%%%%%%%%%%%%%%%
%% Topic distribution in the WhyKB training data.
% 32757 Computers and Electronics
% 25540 Health
% 24411 Food and Entertaining
% 23360 Hobbies and Crafts
% 18332 Home and Garden
% 12929 Education and Communications
% 12913 Personal Care and Style
% 12908 Finance and Business
% 11305 Pets and Animals
% 9145 Arts and Entertainment
% 8743 Youth
% 8040 Sports and Fitness
% 7586 Relationships
% 4864 Family Life
% 4655 Work World
% 3926 Cars & Other Vehicles
% 2162 Holidays and Traditions
% 2133 Travel
% 1897 Philosophy and Religion
%
% Total: 1,759,271 steps from 227,872 procedures (this includes repeated process, about 20% different methods for a process).
% 259,570 train ; 32446 test and 32446 dev. = 324,462

% input = <s1, s2>
% vectorize (s1, s2) => lookup in locality sensitive hash index => finds top-10 neighbors.
% Then aggregates the label distribution of top-10, and this can be used to compute argmax as predicted label.
% ProPara based gold dependencies using heuristics {1: 912, 0: 1289}
% exact match based lookup = 0
% vector-sim match based lookup = xxxx ( / 2201)
% neural match based lookup = 53.5 (1177.0 / 2201 correct points.)  and no entity found in 563 points. Note: tp = 417
% \newpage
\section{Experiments}
\label{sec:evaluation}
% \todo{Switch 4.1 and 4.2: \\
% 4.1:Baselines: Explain that baselines produce $\pi_i_j$ \\
% 4.2:Task: Explain how we can infer enable links from $\pi_i_j$ and hence all baselines can be applied on this task.}

% We report the performance of \ourmodel{} in this section, compared to several other baseline approaches.

%\subsection{Task} \label{sec:task}
\subsection{Evaluation metric and baselines} 
\label{sec:task}

 \finaledit{
  To evaluate \ourmodel, we measure its performance on dependency explanations, as well as its
performance on the original state change prediction task. For state change prediction, we
use the same dataset and evaluation metric described in \cite{propara-emnlp18},
consisting of 1095 test questions for the inputs, outputs, conversions, and movements
in the process (straighforwardly derivable from the predicted state change matrix).
For dependency explanations, we compare the curated, gold graphs $\mathcal{G}_{gold}$ that we added to
ProPara (Section~\ref{dependency-graph-computation}) with the predicted dependency
graphs $\mathcal{G}_{pred}$. Each edge in a graph contains 3 elements of explanation for ``Why $s_t$?'',
namely ``$s_t$ enables $s_u$ due to a state change $\pi_{tk}$ in entity $e_k$'', 
resulting in 2,814 elements to predict in the test set.
We compute the precision and recall of predicting these elements for all steps $s_t$,
yielding an overall $F_1$ score. 
}

% \finaledit{We use the most recent evaluation metric for the state change task \cite{propara-emnlp18} defined over the ProPara dataset. 
% \bhavana{do we need to mention why not the metric from \cite{propara-naacl18}?}
% This metric measures the accuracy of state change predictions but it does not evaluate dependency explanations. 
% To evaluate our proposed auxiliary task of dependency explanations, we 
% generated 2,814 QA pairs by applying the method described in Sec.~\ref{dependency-graph-computation} over ProPara's gold annotations. } 
% An example of this dependency explanation question is ``texttt{which actions does $s_t$ enable and why?}'' with the expected explanation structure ``\texttt{<$s_u$> depends on <$s_t$> due to a state change <$\pi_{tk}$> in entity <$e_k$> at <$s_t$>}''. During evaluation, we report F$_1$ by   comparing the system's predicted dependency graph $\mathcal{G}_{pred}$ with the gold dependency graph $\mathcal{G}_{gold}$ for each paragraph.

%\label{baselines}
We consider the state-of-the-art process comprehension models reported in \cite{propara-emnlp18} as our baselines, including Recurrent Entity Networks (EntNet)~\cite{Henaff2016TrackingTW}, Query Reduction Networks (QRN)~\cite{Seo2017QueryReductionNF}, ProLocal and ProGlobal~\cite{propara-naacl18} and finally, ProStruct~\cite{propara-emnlp18}. These models output state-change predictions, which are used to create the corresponding dependency explanation graph using the method in Section~\ref{dependency-graph-computation}, but do not bias their state change predictions towards those that appear purposeful.
% \finaledit{Also note that \ourmodel{} was developed in parallel with recent KGMRC \cite{Das2018BuildingDK} and NCET \cite{Gupta2019TrackingDA} approaches that establish new state of the art on ProPara dataset. We provide their scores on the dependency task proposed in this paper. }
%We compare our  model \ourmodel{} with these baselines on the original state-change benchmark question set and the proposed question set about dependency graphs. 
% (and can be derived from $\DG_{pred}$ if the system does not produce these explanation explicitly)

% \noindent
% {\bf Rule-based baseline} is similar to \cite{propara-arxiv} that uses rules mapping SRL patterns to state changes, its performance appears limited by the
% incompleteness and approximations in the rulebase, and by errors by the SRL parser. This is additional baseline and not part of the models reported in \cite{propara-emnlp18}.
% \bhavana{Find the prediction file for rule-based-baseline}

\subsection{Results}
\label{results}

\subsubsection{Results on the proposed dependency task:}
% table moved here so it appears first.
\begin{table}[h!]
\centering
\boldmath
\scalebox{0.9}{%
%  \begin{tabular}{|l|ccc|} 
  \begin{tabular}{lccc} 

 \toprule
%  \hline
 Model & P & R & F$_1$ \\
 \toprule
%  \hline
% EntNet  &  26.38  &  53.21  &  35.27 \\
% QRN  &  29.98  &  45.79  &  36.23 \\
% ProLocal  &  92.16  &  27.95  &  42.89 \\
% ProGlobal  &  40.58  &  52.59  &  45.81 \\
% ProStruct  &  89.49  &  33.47  &  48.72 \\
% \ourmodel & 70.70 &	44.80 &	\textbf{54.84}	 \\
% \hline
\newresults{ProLocal} & 24.7	 & 18.0	 & 	20.8	\\
\newresults{QRN} & 32.6	 & 30.3	 &	31.4	\\
\newresults{EntNet} & 32.8 & 	38.6	 & 	35.5	\\

\newresults{ProStruct} & 76.3	 & 21.3	 & 	33.4	\\
\newresults{ProGlobal} & 43.4	 & 37.0	 & 	39.9	\\
% \newresults{\finaledit{NCET}} & 49.63	& 33.76	&	\finaledit{40.18} 	 		\\
% \newresults{\finaledit{KGMRC}} &45.04	 & 45.44	&	\finaledit{45.24} 	 		\\
% \hline
\midrule
\newresults{\ourmodel} & 62.0 &	32.9	&	\textbf{43.0} \\
% \hline
%ProCausal (w/o WhyKB)  &  63.54	& 44.25 &	\textbf{52.17} \\
%ProCausal & 70.90	& 42.76	& \textbf{53.35}	 \\
% \hline
\bottomrule
\end{tabular}
}
\caption{\finaledit{Results on the \textbf{dependency} task (test set). }
%\todo{Add a note: All methods except procausal don't optimize for dependency graph while making $\pi_i_j$ predictions.}
}
\label{table:F1-Why-QA-task}
\vspace{-1mm}
\end{table}
Table~\ref{table:F1-Why-QA-task} reports results of all models on the new \textbf{dependency} task. 
\cameraready{\ourmodel{} significantly outperforms the strongest baselines, ProGlobal and \prostruct{}, by more than 3 points F$_1$}.  \ourmodel{} has much higher precision than ProGlobal with similar recall, suggesting that \ourmodel's dependency-aware decoder helps it select more accurate dependencies. 
% PEC does the below contribute anything? I'd avoid 'causal knowledge', perhaps 'dependency knowledge'? But then it's somewhat vacuuous.
\cameraready{Compared with ProStruct, it yields more than 11.6 points improvement on recall.}
% , suggesting that causal knowledge helps \ourmodel{} discover more dependencies. 
As \ourmodel{} adds a novel dependency layer on top of the ProStruct architecture, we note that all these gains come exclusively from the dependency layer.

% NT: removed the term meaningful dependencies because \ourmodel{} precision is substantially lower.
% NT: this statement was not correct. The quality is _not_ higher in terms of precision.
%This hypothesis is supported by the fact that the best performing baseline, \prostruct{}, yields 7 points lower recall than \ourmodel{} and the quality of the dependencies is higher in terms of precision. 

%-----------------------------------------------------------------------------------

%\subsection{Effects of ProCausal components}
\subsubsection{Impact of \ourmodel{} components:}
\label{ablations}
Table~\ref{table:F1-Why-QA-task-ablation} shows the impact of removing dependency graph scores from \ourmodel. 
Removing $g_{kb}$, results in a substantial (9 points) drop in recall, since the action dependency priors uses background knowledge to help the model discover dependency links that the model could not infer from the current paragraph. 
Further, removing $g_{edge}$ score results in around 2 points drop in recall and a substantial increase in precision (F$_1$ is still lower). % SY: Precision is up for almost 20 points. I think it's better to discuss it as well. 
This is because the $g_{edge}$ score encourages the model to predict state changes that result in more action dependency links, often at the cost of a drop in precision.
Thus, the $g_{edge}$ and $g_{kb}$ scores together help \ourmodel{} discover meaningful action dependency links.
\begin{table}[h]
\centering
\boldmath
\scalebox{0.9}{%
%  \begin{tabular}{|l|ccc|} 
  \begin{tabular}{lccc} 

%  \hline
 \toprule
%  &\multicolumn{3}{|c|}{{Dependency Task (test set)}}  \\
  &\multicolumn{3}{c}{{Dependency Task}}  \\

 Ablation &  P & R & F$_1$ \\
 \toprule
%  \hline 
%  ProStruct & 89.49 & 33.47 &  48.72 \\
%  + $g_{edge}$ & 63.54 & 44.25 & 52.17  \\
%  + $g_{kb}$ & 70.70 & 44.80 &	54.84 \\
%  \ourmodel{} & 70.70 & 44.80 &	54.84 \\
%  - $g_{kb}$ & 63.54 & 44.25 & 52.17 \\
%  - $g_{kb}$ - $g_{edge}$  & 89.49 & 33.47 &  48.72  \\
% \hline
\newresults{\ourmodel{}} & 62.0 &	32.9 &	\textbf{43.0} \\
\newresults{ - $g_{kb}$} & 67.9	& 23.1	&	34.5 \\
\newresults{ - $g_{kb}$ - $g_{edge}$}  & 76.3	 & 21.3	 & 	33.4  \\
% \hline
\bottomrule
\end{tabular}
}
\caption{\finaledit{Effect of ablating $g_{edge}$ and $g_{kb}$ from \ourmodel{} (test set).}
% on the \\ \textbf{dependency} task (test set).
}
% \finaledit{Combine Table 1,2?} => NT: no, I think they are different.
\label{table:F1-Why-QA-task-ablation}
\vspace{-3mm}
\end{table}
% Adding $g_{edge}$ steers the model toward denser \texttt{enables} connections, substantially increasing recall by over 10 points. Simultaneously, there is a drop in precision because many of these newly found \texttt{enables} links may not be sensible. Adding $g_{kb}$ filters such implausible links, boosting precision by 7 points. 

% NT: the following line seems repetitive.
%Overall,  \ourmodel{} has over 12 points relative F1 improvement over prior state-of-the-art, \prostruct, on the dependency task.

%\todo{Add an example highlighting recall and precision gains} \bhavana{Already covered in Section \ref{sec:analysis_procausal_components}}

% \vspace{-2mm}
\subsubsection{Results on the previous state-change task:} \label{sec:results-state-change}

% NT: rephrased the following
%Here we report the results on previous \textbf{state change} task in Table~\ref{table:F1-EMNLP-qset} (all baseline numbers in Table~\ref{table:F1-EMNLP-qset} are from results published. \ourmodel{} maintains high F1 (within 1 point range of the state-of-the-art \prostruct{}. Therefore, \ourmodel{} discovers meaningful dependencies while maintaining good state-change predictions.

\finaledit{On the original state-change prediction task, we find that XPAD performs slightly better (by 0.7 points F1)
than the best published state-of-the-art system ProStruct\footnote{
  Since XPAD was developed, two higher unpublished results of 57.6 and 62.5 on the state-change task have appeared on arXiv \cite{Das2018BuildingDK,Gupta2019TrackingDA}, their systems developed contemporaneously with XPAD.
  In principle XPAD's approach of jointly learning both state changes and dependencies could be also applied in these new systems. Our main contribution
  is to show that jointly learning state changes and dependencies can produce more rational (explainable) results, without damaging (here, slightly improving) the state change predictions themselves.},
even though Equation 1 is not optimized solely for that task (Table~\ref{table:F1-EMNLP-qset}).
This illustrates that encouraging purposefulness in action prediction not only produces
more explainable action sequences, but can improve the accuracy of those action predictions themselves.}

\begin{table}[!h]
\centering
\boldmath
%\scalebox{0.7}{%
%  \begin{tabular}{|l|ccc|} 
  \begin{tabular}{lccc} 
%  \hline
 \toprule
 & P & R & F$_1$ \\
%  \hline
 \toprule
% EntNet  &  54.7  &  30.7  &  39.4 \\
% QRN  &  60.9  &  31.1  &  41.1 \\
% ProLocal  &  81.7  &  36.8  &  50.7 \\
% ProGlobal  &  61.7  &  44.8  &  51.9 \\
%ProCausal (w/o WhyKB)  &  68.9	& 43.3	& 53.2 \\
% ProCausal & 71.7 &	44.7 &	\textbf{55.0}  \\

ProStruct  &  74.3  &  43.0  &  54.5 \\
\ourmodel & 70.5 &	45.3 &	\textbf{55.2}  \\
% \finaledit{- dependency layer} & 74.3  &  43.0  &  54.5  \\
\bottomrule
% \hline
\end{tabular}
%}
\caption{\finaledit{Results on the \textbf{state-change} task (test set), comparing with the best published prior result.} 
% ProStruct is the prior state-of-the-art model.
}
\label{table:F1-EMNLP-qset}
\vspace{-2mm}
\end{table}

% \begin{table}[tbh]
% \centering
% \boldmath
% \scalebox{0.9}{%
%  \begin{tabular}{|l|ccc|} 
%  \hline
%  &\multicolumn{3}{|c|}{{Dependency Task}}  \\
%  & $\triangle$ P & $\triangle$R & $\triangle$F1 \\
%  \hline
% ProCausal &  &  &   \\
%  - no L_{graph} & -9.1 & 0.44 & -2.45 \\
%  - no L_{KB} & -4.6 & 0.64 & -0.85 \\
% \hline
% \end{tabular}
% }
% \caption{Ablation results for ProCausal model on dependency task (w.r.t. test set). }
% \label{table:F1-Why-QA-task-ablation}
% \vspace{-2mm}
% \end{table}

% \begin{table}[tbh]
% \centering
% \boldmath
% \scalebox{0.9}{%
%  \begin{tabular}{|l|ccc|ccc|} 
%  \hline
%  &\multicolumn{3}{|c|}{{Dependency Task}} &\multicolumn{3}{|c|}{{State change Task}} \\
%  & $\triangle$ P & $\triangle$R & $\triangle$F1 & $\triangle$ P & $\triangle$R & $\triangle$F1 \\
%  \hline
% ProCausal &  &  &  &  &  &  \\
%  - no L_{graph} & -9.1 & 0.44 & -2.45 & -5.9 & -8.0 & -7.9 \\
%  - no L_{KB} & -4.6 & 0.64 & -0.85 & 0.6 & -4.1 & -3.0 \\
% \hline
% \end{tabular}
% }
% \caption{Ablation on both the tasks (w.r.t. test set). }
% \label{table:F1-Why-QA-task-ablation}
% \vspace{-2mm}
% \end{table}

\section{Analysis and Discussion} \label{sec:discussion_analysis}
\vspace{-2mm}

\subsection{Dependency Graph Computation Errors}
\label{dependency-graph-errors}
\finaledit{Our heuristic algorithm for estimating the dependency graph from state change
predictions (Section~\ref{dependency-graph-computation}) makes a {\it coherence assumption},
namely that the purpose of a state change on entity $e$ is to enable the next
event mentioning $e$. We now describe classes of errors that the
algorithm makes, by summarizing errors that had to be corrected when
curating the gold explanation graphs for the test set. 
} \\
\noindent
{\bf 1. Coreference (synonymy)} (35\% of errors):
Reference to a changed entity is missed if an alternative wording is used, e.g.,
\begin{myquote}
{\it snow becomes {\bf ice}...{\bf the mass} grows smaller...}
\end{myquote}

\noindent 
{\bf 2. Unstated Linkage} (25\%):
Sometimes use of an entity will be implicit, e.g.,
\begin{myquote}
{\it Animals eat plants...Animals make waste {\bf [from the plants]}...}
\end{myquote}
preventing finding what ``eat plants'' enables.

\noindent 
{\bf 3. Ellipsis} (15\%):
(Sometimes ungrammatical) elided references to earlier entities will hide the purpose of those entity's state changes, e.g.,
\begin{myquote}
{\it {\bf Water} flows downwards... Enters the dam...}
\end{myquote}
Thus, this analysis suggests several ways that the dependency
graph computation could be improved.

\noindent
{\bf 4. Bridging Anaphora} (15\%):
In some cases, a sentence will refer indirectly to a changed entity, e.g.,
by referring to its part, whole, or some other association (i.e., an
associative/bridging anaphoric reference \cite{Wei2014ASO}). 
As the algorithm does not resolve such indirect references, the
purpose of the action changing that entity is unrecognized, e.g.,
\begin{myquote}
{\it spark plug causes a {\bf spark}...{\bf explosion} occurs...} \\
{\it The {\bf leaf} absorbs CO$_2$... The {\bf plant} produces O2..}
\end{myquote}

% \vspace{1mm}
\noindent
{\bf 5. Bad State Change Annotations} (5\%):
In a few cases, state change annotations are missing or in error in the gold dataset,
resulting in missing or incorrect prediction of their purpose.

\noindent {\bf 6. Long-Range Dependencies} (5\%):
We assume that an entity's change is to enable the next mentioned use of that
entity. This assumption is based solely on typical protocols of discourse,
and occasionally is violated. For example:
\begin{myquote}
{\it Rainwater picks up CO$_2$... Rainwater goes into the soil... The water dissolves limestone...}
\end{myquote}
Here the first action is a prerequisite for the third, not the second,
thus violating our assumption and creating an error in the dependency graph.

Note that around 45\% of the sentences in \dataset~have no associated
state change because of unmodeled state changes (other than create/destroy/move) or the presence of stative sentences (no resultant state changes). As a result, the purpose of such sentences is inherently unrecoverable given these annotations, and out of reach of \ourmodel~(and any other model).

\subsection{Qualitative Analysis of \ourmodel's Results}

\subsubsection{Corrected Predictions}
\label{sec:analysis_procausal_components}
In many cases, the bias toward \texttt{enables} links results in \ourmodel~predicting
a state change when previously \statechange{none} had been predicted. For
example, in sentence (6) of the paragraph snippet below:
\begin{myquote}
{\it (6) Millions of years later the fossil forms.} \\
{\it (7) A person finds the fossil.} \\ 
{\bf \prostruct~(for step 6):} \statechange{None} \\
{\bf \ourmodel:} \statechange{CREATE(fossil)} {\bf [correct]}
\end{myquote}
\prostruct~incorrectly predicted no state changes from (6), while
\ourmodel~(correctly) predicted that a fossil was created. The
extra evidence causing \ourmodel~to make this prediction is that
it adds an edge in the dependency graph, giving step (6) a purpose,
namely to enable step (7) in which the fossil is found. There
are many such examples.

\subsubsection{Erroneous Predictions}

Both of \ourmodel's new biases encourage \ourmodel~to make state-change predictions
that result in more \texttt{enables} edges, either by counting them ($g_{edge}$)
or summing their likelihoods ($g_{kb}$), resulting in some overlap
in their overall effect (Table~\ref{ablations}). The bias can help
predict missing state changes (e.g., above), but can also
cause \ourmodel~to ``hallucinate'' state changes with weak
evidence, simply to give each sentence a purpose.
An example is:
\begin{myquote}
  {\it (5) Carbon-based mixture stays underground.} \\
  {\it (6) Humans discover this carbon-based mixture.} \\
  {\bf \prostruct~(for step 5):} None {\bf [correct]} \\
  {\bf \ourmodel:} CREATE (carbon-based mixture)
\end{myquote}
Here, there is no action associated with step (5). However,
in part because this is the first mention of carbon-based mixture, % in the text,
\ourmodel~incorrectly labels it as being created, giving step (5)
a purpose of enabling step (6). Although such errors occur, they
are less common than the corrected predictions, hence the
improved overall performance. % in Table~\ref{results}.

Comparing the ablated versions of \ourmodel, we can also see how the
different biases play out. Without $g_{kb}$, \ourmodel~can sometimes
predict \texttt{enables} edges that are nonsensical with respect to background
knowledge, for example, 
\begin{myquote}
  {\it (1) Nitrogen exists naturally in the atmosphere.} \\
  %{\it $\vdots$} \\
  {\it (9) Bacteria turn nitrogen into a gas.}  \\
{\bf \ourmodel~(no $g_{kb}$) (for step 1):} CREATE (gas)  \\
{\bf \ourmodel:} None {\bf [correct]}
\end{myquote}
In this example, \ourmodel~without $g_{kb}$ produces an incorrect %, strange
prediction for (1) (that \entity{gas} has been created), as it results
in an \texttt{enables} link from step (1) to (9). However, the background KB deems
it unlikely that (1) enables (9); as a result, the full \ourmodel~makes 
no such prediction.

% \vspace{-1mm}
\section{Conclusion}
% \vspace{-1mm}
Our goal is to better comprehend procedural text by not only predicting
{\it what} happens, but {\it why}. % some events need to happen before others. 
To do this, we have expanded the
traditional state-tracking task with the additional task of predicting and explaining
dependencies between steps, and presented a new model, \ourmodel, for
these tasks. \ourmodel~is biased to prefer predictions that have an
identifiable purpose (enables a subsequent step),
and where that purpose is more plausible (judged using a large corpus).
Experiments show that \ourmodel~significantly improves the predictions
of correct dependencies between actions, while matching state-of-the-art
results on the earlier tasks.

Although our experiments have been with ProPara, there is nothing intrinsic
to \ourmodel's architecture that limits it to that dataset's design:
Given appropriate annotations, \ourmodel~could be trained with
a richer set of state change operators, longer paragraphs, 
and (with the addition of a temporal ordering module,
e.g., \cite{ning2017structured}) non-chronological event orderings. 
\cameraready{In addition, given additional hand-labeled training data, a system might learn to directly predict dependency links (instead of deriving them heuristically from state changes).
These would be valuable new directions to pursue.
The Dependency Graph dataset is available within 
 \small \url{http://data.allenai.org/propara}}

\subsection*{Acknowledgements}
We are grateful to the AllenNLP and Beaker teams at AI2, and for the insightful discussions with other Aristo team members. 
Computations on beaker.org were supported in part by credits from Google Cloud.

\bibliography{references}
\bibliographystyle{acl_natbib}

\end{document}